\documentclass[letterpaper,10pt,journal,twoside]{ieeetran}

\IEEEoverridecommandlockouts                              % This command is only needed if 
\usepackage{times}
\usepackage[pdftex]{graphicx}
\usepackage{amsmath,amssymb,amsopn,amstext,amsfonts}
\usepackage{cancel}
\usepackage[space]{cite}
\usepackage{color}
\usepackage{mathtools}
\usepackage{bm}

\usepackage{diagbox}
\usepackage{float}
\usepackage{epstopdf}
\usepackage{pifont}
\usepackage{fixltx2e}
\usepackage{amsmath}
\usepackage{multirow}
\usepackage{url}
\usepackage{verbatim}
\usepackage{soul,xcolor}
\usepackage[linesnumbered,ruled,vlined]{algorithm2e}
\usepackage{subfiles}
\usepackage{array}
\usepackage{hhline}
\usepackage{makecell}
\usepackage{booktabs}

\usepackage{diagbox}
\usepackage{threeparttable} 
\graphicspath{{./img_compress/}{../img_compress/}}
\DeclareGraphicsExtensions{.png,.jpg,.eps,.pdf,.jpeg}
\makeatletter
\let\NAT@parse\undefined
\makeatother
\usepackage[colorlinks=true, linkcolor=black, anchorcolor=black, citecolor=black, urlcolor=black, citebordercolor=black]{hyperref}
\hypersetup{
    pdftitle={Multi-Floor Exploration for Ground Robots via an Incremental Reachable Graph and Structural Priors},
    pdfauthor={Zhiwen Zhu, Jiaqi Chen, Xiangyi Huang, Meiqi Hu, Boyu Zhou}
}

\makeatletter
\def\section{\@startsection{section}{1}{\z@}{0.48ex plus 0.18ex minus 0.1ex}%
{0.36ex plus 0.15ex minus 0ex}{\normalfont\normalsize\centering\scshape}}%
\def\subsection{\@startsection{subsection}{2}{\z@}{0.37ex plus 0.15ex minus 0.1ex}%
{0.26ex plus 0.1ex minus 0ex}{\normalfont\normalsize\itshape}}%
\def\subsubsection{\@startsection{subsubsection}{3}{\z@}{0.27ex plus 0.1ex minus 0.1ex}%
{0.16ex plus 0.1ex minus 0ex}{\normalfont\normalsize\itshape}}%
\makeatother

\makeatletter
\def\@cite#1#2{{\color{black}[{#1\if@tempswa , #2\fi}]}}
\makeatother

\makeatletter
\let\@oldmaketitle\@maketitle% Store \@maketitle
\renewcommand{\@maketitle}{\@oldmaketitle% Update \@maketitle to insert the teaser figure.
    \centering
    \setcounter{figure}{0}
    \includegraphics[width=2.0\columnwidth]{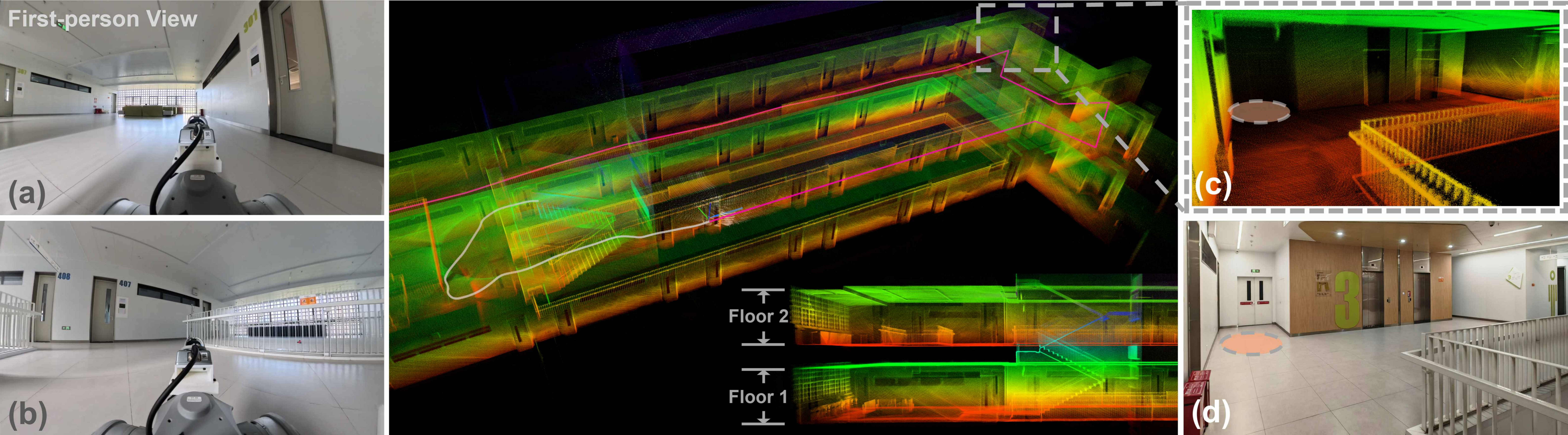}
    \vspace{0.2cm}
    \refstepcounter{figure}\label{fig:head_img}
    \parbox{2.0\columnwidth}{\footnotesize Fig.~\thefigure.\enspace
    Real-world multi-floor exploration in a campus building. The central image depicts an exploration snapshot on the second floor, visualizing the quadruped robot's odometry, trajectory, current point cloud (white), final reconstructed map (colored), and global guidance (purple) across the floor. 
    Images (a-b) compare scene photographs at vertically corresponding locations, revealing distinct local layouts across different floors. 
    Images (c-d) illustrate a small region (orange) that is easily overlooked by conventional exploration relying solely on observed maps, which often requires inefficient revisits. In contrast, the proposed method leverages projected structural priors to guide the planner, enabling coverage during the initial pass.
    }
    \vspace{-1.0cm}
}
\makeatother

\title{\LARGE \bf
Multi-Floor Exploration for Ground Robots via an Incremental Reachable Graph and Structural Priors
}

\author{Zhiwen Zhu, Jiaqi Chen, Xiangyi Huang, Meiqi Hu, Boyu Zhou}

\begin{document}

\maketitle
\thispagestyle{empty}
\pagestyle{empty}

%%%%%%%%%%%%%%%%%%%%%%%%%%%%%%%%%%%%%%%%%%%%%%%%%%%%%%%%%%%%%%%%%%%%%%%%%%%%%%%%

\begin{abstract}
Autonomous exploration of multi-floor buildings remains challenging for ground robots because conventional 2D and 2.5D maps cannot represent overlapping traversable surfaces such as stairs, ramps, and multiple reachable elevations. This letter presents a multi-floor exploration framework based on an incremental reachable graph. Built as a sparse graph over reachable support surfaces, the graph preserves potentially valid connectivity through tentative graph elements under sparse observations and enables stable, physically reachable frontier detection. To guide exploration beyond the currently mapped floor, we project task-zone priors from an explored floor to initialize a hypothetical graph on the target floor and reconcile it incrementally with incoming observations. A hierarchical planner then jointly reasons over confirmed and hypothetical structures for global guidance. In simulation, the proposed method demonstrates improved exploration efficiency and mapping completeness compared to evaluated baselines. Furthermore, onboard real-world experiments validate its practical feasibility and real-time performance.
\end{abstract}

%%%%%%%%%%%%%%%%%%%%%%%%%%%%%%%%%%%%%%%%%%%%%%%%%%%%%%%%%%%%%%%%%%%%%%%%%%%%%%%%

%%%%%%%%%%%%%%%%%%%%%%%%%%%%%%%%%%%%%%%%%%%%%%%%%%%%%%%%%%%%%%%%%%%%%%%%%%%%%%%%


\section{Introduction}
\label{sec:intro}

Autonomous exploration and mapping in building environments are critical for applications such as search and rescue, structural inspection, and digital twin construction. Ground robots, including legged and wheeled platforms, offer higher payload capacity and longer endurance than aerial vehicles, making them attractive for sustained exploration. However, most existing exploration methods for ground robots rely on 2D occupancy maps or elevation maps, implicitly assuming that the navigable space can be represented by a single horizontal layer or a single-valued height field. This assumption breaks down in buildings with stairs, ramps, and multi-floor layouts, where a single planar location may correspond to multiple traversable heights. As a result, conventional representations fail to capture the true navigable structure and limit exploration to approximately planar environments.

To overcome this in multi-elevation environment, the representation must explicitly model 3D traversable terrain. Unlike aerial robots navigating in unconstrained free space, ground robots require continuous terrain support, making exploration reliant on this structural connectivity. In practice, onboard LiDAR sensors suffer from limited mounting height and inherent scanning characteristics, yielding dense, reliable observations only at close range. For distant regions, near-ground measurements are often sparse and fragmented. Consequently, restricting planning to only fully verified traversable areas leads to excessive conservatism: the robot may not explore incompletely observed boundaries, overlooking potentially valid traversable connections and narrowing the effective exploration scope. This raises our first key challenge: maintaining an environmental representation that better captures potentially explorable space under incomplete and fragmented observations.

To address this issue, we model the navigable space with an incremental reachable graph. Specifically, we construct a sparse graph over an online traversability map, preserving unverified but potentially valid connectivity as tentative elements in under-observed yet obstacle-free regions. As observations accumulate, these elements are confirmed or pruned, and disconnected components are removed to maintain topological consistency. For ground robots, frontiers extracted from volumetric free-unknown boundaries often yield invalid targets, such as mid-air spaces or isolated regions disconnected from navigable terrain. Likewise, relying on reconstructed surface boundaries typically produces fragmented and temporally inconsistent targets. We therefore define frontiers directly on the maintained graph. A node is designated as a frontier if it indicates local surface incompleteness or topological extensibility into unknown space, ensuring that all exploration targets remain physically reachable and structurally stable.

A second challenge arises in multi-floor exploration. Most existing methods plan only within the currently mapped region and therefore provide little guidance toward yet-unobserved floors. In real buildings, however, adjacent floors often exhibit coarse structural regularities, such as aligned corridors, stairwells, and load-bearing walls. These regularities can provide useful guidance even before the next floor is sufficiently observed. Motivated by this observation, we transfer structural priors from an explored floor to an unexplored one. We partition the maintained graph of the explored floor into macroscopic task zones and project them onto the target floor to initialize a hypothetical graph. As exploration proceeds, this hypothesis is incrementally reconciled with actual observations by preserving valid priors and correcting inconsistent ones. A hierarchical planner then jointly reasons over confirmed and hypothetical structures, enabling global guidance beyond the currently mapped region and improving exploration efficiency across the building.
The main contributions are as follows:

1) An incremental reachable graph that preserves potentially valid but unconfirmed connectivity through tentative graph elements, enabling stable and physically reachable frontier discovery for ground robots.

2) A cross-floor structural prior transfer mechanism that initializes and incrementally reconciles a hypothetical graph on unexplored floors, enabling global exploration guidance before sufficient observations are available.

\section{Related Work}
\label{sec:related}

\subsection{Traversability-Aware Mapping for Ground Robots}
Mapping and planning methods for ground robots commonly rely on 2D occupancy grids or 2.5D elevation maps. Occupancy grids provide a simple, efficient model for near-planar environments~\cite{elfes1989using,zhang2025ddr}, while elevation maps estimate the support surface and its uncertainty in rough terrain~\cite{miki2022elevation}. Recent studies use Bayesian inference or Gaussian processes to infer traversability and risk from local point clouds~\cite{oh2024trip,hou2025fsgp}. While effective for local terrain assessment, these methods typically represent each planar location with only one support height or traversability estimate. They therefore struggle in indoor multi-elevation environments with stairs, overhangs, and mezzanines, where the same planar location may correspond to multiple traversable elevations.

To better capture 3D structure, richer representations have been explored, including point clouds, meshes, plane abstractions, and tomogram slices~\cite{krusi2017driving,brandao2020gaitmesh,jian2022putn,xu2024pare,zhang2025planes,yang2025tomography}. However, most such methods target goal-directed navigation on prebuilt or locally reconstructed maps rather than online autonomous exploration. Inspired by the terrain feature extraction in~\cite{li2025multilevel}, we classify online-mapped 3D voxels into traversable and non-traversable states for exploration. Sparse graph abstractions have also proven effective for improving planning efficiency in large-scale scenes~\cite{lee2025trg}. Unlike existing sparse-graph methods that mainly accelerate planning over already observed traversable regions, our graph is incrementally maintained over reachable support surfaces and preserves tentative yet physically plausible connectivity under incomplete observations.

\subsection{Autonomous Exploration for Ground Robots}
Autonomous exploration iteratively selects observation poses to expand the mapped area. Classical frontier-based methods drive the robot toward boundaries between known and unknown space~\cite{yamauchi1998frontier}, while next-best-view approaches rank candidates by expected information gain~\cite{bircher2016receding}. Later works improved scalability through incremental frontier management and hierarchical planning~\cite{zhou2021fuel,cao2021tare}. Coverage-guided strategies~\cite{zhang2024falcon,zhou2023racer} assume continuously traversable free space and compute global coverage paths to mitigate the limitations of planning on partially observed maps, but this assumption rarely holds in building environments with occlusions and disconnected structures. Moreover, most of these methods were designed for aerial robots and extract exploration targets from volumetric free-unknown boundaries. For ground robots, such targets lack physical reachability guarantees and often correspond to suspended or inaccessible regions.

Existing ground exploration methods differ primarily in target formulation and environment representation. Some plan on 2D grids or projected maps~\cite{liu2024sfre}, which inevitably collapse multi-elevation structures and obscure cross-floor connectivity. To handle uneven terrain, other strategies extract viewpoints directly from reconstructed surfaces or planar features~\cite{wang2020activemapping,azpurua2021terrain,xu2024pare}; however, sparse ground LiDAR measurements often make such targets scattered and unstable. Alternatively, free-voxel frontiers designed for aerial robots have been adapted using heuristic filtering~\cite{huang2023fael,cao2021tare}, though this often compromises map completeness, for example, by neglecting ground-surface modeling. Rather than relying on raw voxel boundaries or surface fragments, our method defines frontiers directly on a 3D traversability-aware reachable graph, thereby preserving physical reachability, maintaining structural completeness, and reducing computational overhead.

When the environment spans multiple floors, existing techniques either decouple each floor into an independent 2D mapping task~\cite{chen2025lite} or simply exhaust all frontiers on the current floor before transitioning~\cite{jia2025multiresolution}. As a result, cross-floor reasoning remains weakly integrated into online exploration. Our framework instead projects task-zone structure from an explored floor to initialize a hypothetical graph on the next, enabling earlier target-floor guidance through structural priors before that floor is fully observed.

\begin{figure*}[t]
	\begin{center}
        % \vspace{0.3cm}
      %\includegraphics[width=2.0\columnwidth]{img_compress/pipline_low.png}     
      \includegraphics[width=1.97\columnwidth]{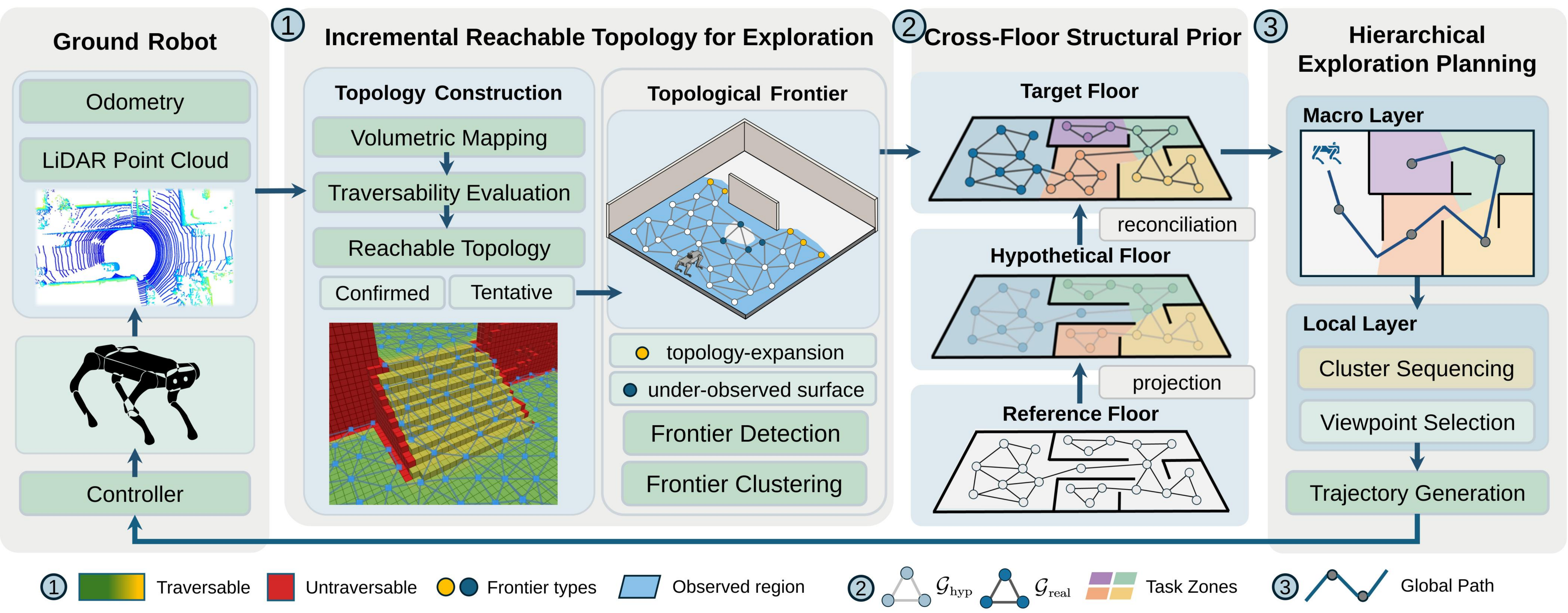}     
      % \vspace{-0.9cm}
	\end{center}
    \vspace{-0.3cm}
   \caption{\label{fig:pipeline} 
   System architecture of the proposed multi-floor exploration framework. Module 1 maintains an incremental reachable graph for exploration by constructing a traversability-aware volumetric map, extracting reachable support structure, and preserving potentially valid connectivity through confirmed and tentative graph elements. Topological frontiers are then defined and clustered on this maintained graph, where orange and dark blue nodes denote frontier nodes triggered by the under-observed surface and topology-expansion criteria, respectively. Module 2 exploits cross-floor structural priors by projecting the task-zone partition ($\mathcal{Z}_{\mathrm{ref}}$) from a reference floor to initialize a hypothetical graph ($\mathcal{G}_{\mathrm{hyp}}$) on the target floor. This hypothesis is incrementally maintained via node promotion ($\mathcal{S}_{\mathrm{hyp}}\rightarrow\mathcal{S}_{\mathrm{conf}}$) and dynamic task-zone reassignment based on actual observations. Module 3 performs hierarchical planning, generating a global sequence across the hybrid graph and locally selecting viewpoints for efficient execution.
   }
  %  \vspace{-0.3cm}
\end{figure*}

\section{Traversability-Aware Incremental Reachable Graph and Topological Frontiers}
\label{sec:representation_tf}

Ground robots must move on physically supported and reachable surfaces. Rather than planning in generic free space, we incrementally model the reachable support structure to guide exploration under incomplete observations. To this end, we maintain a traversability-aware volumetric map (Sec.~\ref{subsec:trav_map}) together with an evolving reachable graph (Sec.~\ref{subsec:sparse_graph}). This graph preserves potentially valid connectivity through tentative elements and supports frontier discovery and viewpoint generation (Sec.~\ref{subsec:tf}), enabling scalable exploration in complex multi-elevation environments.

\subsection{Traversability Map Construction}
\label{subsec:trav_map}

We maintain a volumetric occupancy map updated incrementally via raycasting. For each occupied voxel~$v$, point cloud statistics, including the mean position and covariance matrix, are accumulated online. To evaluate local terrain quality, these statistics are aggregated within a spatial neighborhood of~$v$, and eigendecomposition is performed on the resulting covariance matrix~\cite{li2025multilevel}. Let $\lambda_1 \le \lambda_2 \le \lambda_3$ denote the eigenvalues and $\mathbf{n}_v$ the eigenvector corresponding to~$\lambda_1$ (the estimated surface normal). The local slope $s(v)$ and roughness $r(v)$ are defined as
\begin{equation}
s(v) = \arccos\!\bigl(|\mathbf{n}_v^\top \mathbf{e}_z|\bigr), \quad
r(v) = 1 - \frac{\lambda_2 - \lambda_1}{\lambda_2 + \lambda_1},
\end{equation}
where $\mathbf{e}_z$ is the gravity-aligned unit vector. The under-observation level $u(v)$ measures the sparsity of accumulated points. The composite terrain cost is
\begin{equation}
c(v) = w_s \bar{s}(v) + w_r \bar{r}(v) + w_u \bar{u}(v).
\end{equation}

Each voxel is assigned a traversability state~$\tau(v)$ according to its point count~$N(v)$ and two safety predicates: collision~$\mathcal{R}_{\mathrm{col}}(v)$ (occupied voxels exist above~$v$ within a safety clearance) and fall-risk~$\mathcal{R}_{\mathrm{fall}}(v)$ (insufficient support below~$v$):
\begin{equation}
\label{eq:trav_condition}
\begingroup
\renewcommand{\arraystretch}{0.85}
\tau(v) =
\begin{cases}
\mathcal{S}_{\mathrm{unobs}}, & N(v) < N_{\min}, \\
\mathcal{S}_{\mathrm{trav}}, & c(v) \le \tau_c,\ \neg\mathcal{R}_{\mathrm{col}}(v),\ \neg\mathcal{R}_{\mathrm{fall}}(v), \\
\mathcal{S}_{\mathrm{untrav}}, & \text{otherwise.}
\end{cases}
\endgroup
\end{equation}
Here, $\mathcal{S}_{\mathrm{unobs}}$ denotes an under-observed voxel, and $\tau_c$ is the maximum terrain cost threshold.

Local traversability does not imply global reachability: isolated surfaces such as tabletops may satisfy the geometric criteria while remaining disconnected from the robot. We therefore extract a connected reachable set $\mathcal{C}_{\mathrm{reach}}$ by region growing from the robot's current support voxel over all traversable voxels. This set defines the currently confirmed support structure on which the reachable graph is maintained.

\subsection{Incremental Reachable Graph Construction}
\label{subsec:sparse_graph}

To enable scalable planning in extensive scenes, rather than simply expanding a sparse graph over traversable surfaces, we maintain a sparse reachable graph that represents the topology of the reachable support structure under incomplete support-surface observations, denoted as the real graph $\mathcal{G}_{\mathrm{real}} = (\mathcal{V}_{\mathrm{real}}, \mathcal{E}_{\mathrm{real}})$. Each node $n_i \in \mathcal{V}_{\mathrm{real}}$ stores a support position $\mathbf{p}_i \in \mathbb{R}^3$ matching the local terrain, and each edge $e_{ij} \in \mathcal{E}_{\mathrm{real}}$ represents a collision-free transition.

\noindent\textbf{Spatial expansion.}\; The graph expands by sampling candidate positions around existing nodes. Given a node~$n_i$, a candidate is generated as
\begin{equation}
\mathbf{q} = \mathbf{p}_i + \bigl[r_e\cos\theta,\; r_e\sin\theta,\; 0\bigr]^\top, \quad \theta \sim \mathcal{U}(0,2\pi),
\end{equation}
where $r_e$ is the expansion radius. The candidate is projected onto the local ground surface and accepted only if the underlying voxel is traversable. If an existing node already lies within a minimum clearance of the candidate, no new node is created; instead, an edge is directly established to the existing neighbor. Otherwise, a new node is instantiated and connected to both the reference node and other reachable neighbors. To maintain sparsity, each node is constrained to a maximum degree~$k_{\max}$, and a new edge is accepted only if its direction differs from every existing edge at the same endpoint by at least~$\alpha_{\min}$.

\begin{figure}[t]
    \begin{center}
    \includegraphics[width=\linewidth]{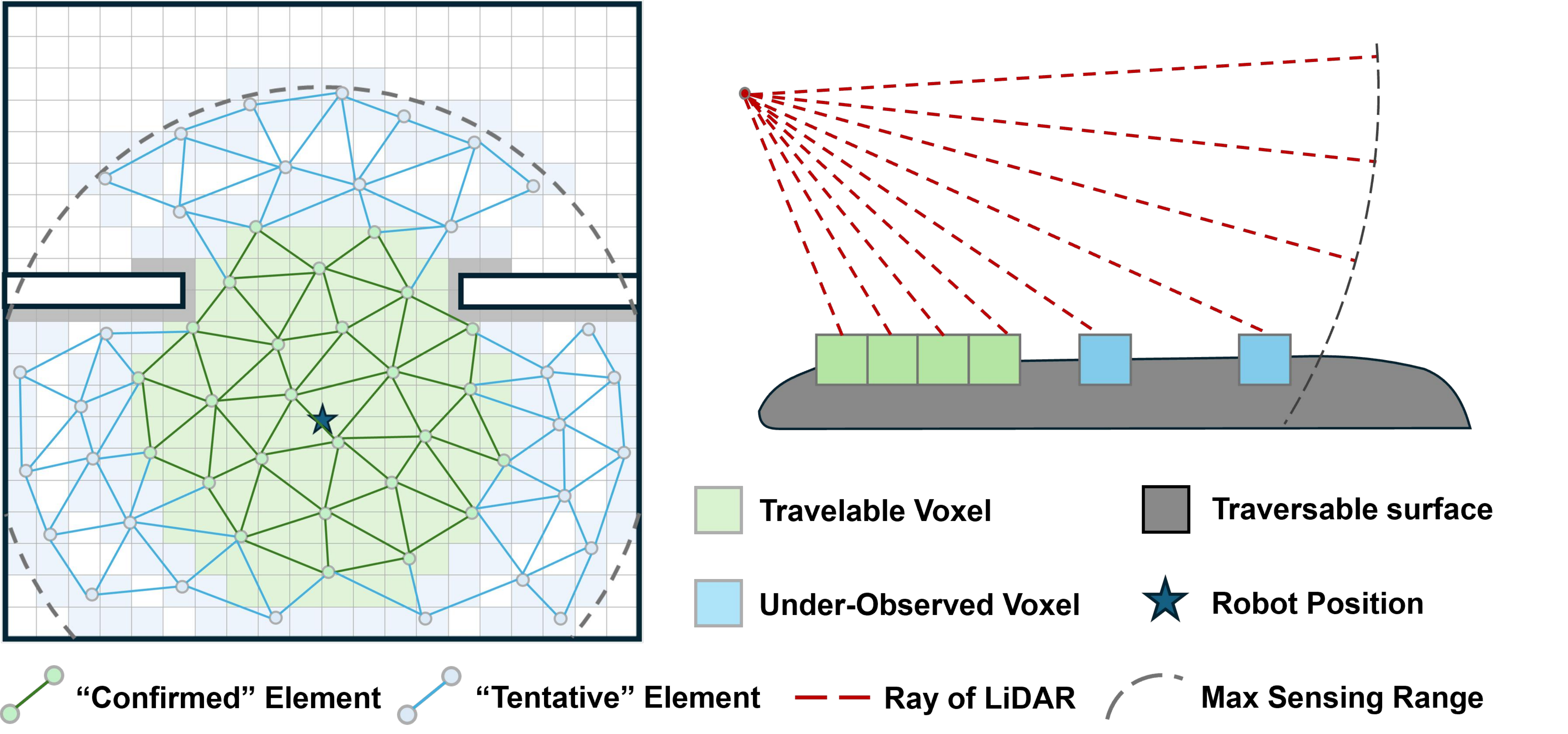}
    \end{center}
    \vspace{-0.3cm}
    \caption{\label{fig:tentative_graph}
    A 2D example illustrating how tentative nodes preserve potential connectivity across under-observed gaps. Near the sensing-range boundary, observations are often incomplete; without tentative elements (blue), the graph would remain confined to the limited confirmed region (green).
    }
\end{figure}

\noindent\textbf{Node state assignment.}\; Ground surfaces near observation boundaries are often partially observed and fragmented. If the graph is restricted strictly to $\mathcal{C}_{\mathrm{reach}}$, it may stop at these under-observed regions and discard potentially valid traversable connectivity, thereby narrowing the exploration scope and delaying the discovery of feasible connecting routes. We therefore assign each element one of two states: \emph{confirmed}~($\mathcal{S}_{\mathrm{conf}}$) if fully supported by~$\mathcal{C}_{\mathrm{reach}}$, or \emph{tentative}~($\mathcal{S}_{\mathrm{tent}}$) if it is collision-free and free of fall risk ($\neg\mathcal{R}_{\mathrm{col}} \wedge \neg\mathcal{R}_{\mathrm{fall}}$). As illustrated in Fig.~\ref{fig:tentative_graph}, tentative elements preserve such potentially valid links, allowing the graph to bridge under-observed regions before they are fully confirmed.

\noindent\textbf{Incremental graph update.}\; Upon each map update, the affected local subgraph is re-evaluated. Tentative elements covered by~$\mathcal{C}_{\mathrm{reach}}$ are promoted to confirmed, while those intersecting newly detected obstacles are pruned. If structural connectivity changes, a breadth-first search rooted at the robot removes isolated components, ensuring that $\mathcal{G}_{\mathrm{real}}$ remains consistent with the actual reachable support structure. The resulting graph can then be organized into macroscopic \emph{task zones} for higher-level planning (Sec.~\ref{subsec:taskzone}).

\subsection{Topological Frontier Definition and Clustering}
\label{subsec:tf}

The maintained reachable graph provides a suitable substrate for frontier discovery. Standard volumetric or reconstructed boundary methods often yield unreachable or fragmented frontier candidates. We address these issues by defining \emph{topological frontiers} (TFs) directly on the nodes of $\mathcal{G}_{\mathrm{real}}$, bypassing raw voxel-level adjacency entirely.

A node $n_i \in \mathcal{V}_{\mathrm{real}}$ is designated as a TF based on two complementary criteria: $C_{\mathrm{sur}}$ captures support surfaces that are not yet fully observed, while $C_{\mathrm{exp}}$ captures locations where the graph may extend into new reachable regions.

\noindent\textbf{1) Under-observed surface criterion.}\; The reconstructed ground surface may contain under-observed gaps or incomplete boundaries. For a node~$n_i$ at height~$z_i$, we define a local horizontal region $\Omega_i$ with radius proportional to the graph expansion distance~$r_e$ and compute the missing-surface ratio:
\begin{equation}
\rho_i = 1 - \frac{1}{|\Omega_i|} \sum_{\mathbf{x} \in \Omega_i} \mathbb{I}(\mathbf{x}, z_i),
\end{equation}
where $\mathbf{x}$ is a sampled horizontal position in $\Omega_i$, and $\mathbb{I}(\mathbf{x}, z_i)$ indicates whether a mapped support surface exists near $(\mathbf{x}, z_i)$. The criterion $C_{\mathrm{sur}}(n_i)$ is satisfied if $\rho_i \ge \rho_{\min}$.

\noindent\textbf{2) Topology-expansion criterion.}\; This criterion identifies nodes from which the reachable graph can extend its connectivity into unknown space. For node~$n_i$, radial probes are cast along directions not covered by existing edges. A direction is considered open if it traverses known free space and reaches unknown regions without intersecting obstacles. Let $\Theta_i^{\mathrm{open}}$ denote the set of such open directions. The criterion $C_{\mathrm{exp}}(n_i)$ is satisfied if $|\Theta_i^{\mathrm{open}}| \ge \Theta_{\min}$.

A node is classified as a frontier if either condition is met, i.e., $n_i \in \mathcal{V}_{\mathrm{TF}} \iff C_{\mathrm{sur}}(n_i) \lor C_{\mathrm{exp}}(n_i)$.

Because both criteria operate on graph nodes rather than raw voxel boundaries, frontier candidates are restricted to the maintained reachable support structure. This inherently enforces physical reachability, avoids suspended or floating artifacts, and reduces the computational state space to the sparse graph.

Individual TF nodes are grouped into clusters by iteratively merging connected neighbors along graph edges up to a maximum capacity. By relying on topological connectivity rather than Euclidean proximity, this strategy prevents the erroneous merging of frontiers separated by structural barriers such as walls. The spatial centroid of each cluster serves as its representative location.

Candidate viewpoints are sampled from nearby graph nodes and assigned a yaw angle facing the cluster centroid. To evaluate viewpoint quality, we simulate raycasting from each candidate to estimate the number of visible TF nodes while explicitly accounting for sensor field-of-view limits and volumetric occlusions. Candidates with high expected visibility are retained for subsequent exploration planning (Sec.~\ref{sec:multilayer_planning}).

\section{Hierarchical Exploration Planning with Structural Priors}
\label{sec:multilayer_planning}

Given the TF clusters and their candidate viewpoints from Sec.~\ref{sec:representation_tf}, the exploration planner must determine their visiting order and select one viewpoint per cluster. We employ a two-layer architecture. A \emph{macro layer} organizes the sparse graph into task zones for global layout-level guidance (Secs.~\ref{subsec:taskzone}--\ref{sec:global_planning}), while a \emph{local layer} solves a Traveling Salesman Problem (TSP) over TF clusters and selects viewpoints through a heading-aware dynamic program (Sec.~\ref{sec:local_trajectory}).

In multi-floor environments, vertically adjacent floors typically exhibit strong structural regularities, such as aligned corridors and load-bearing walls. Once one floor is explored, the macro layer projects its task-zone layout onto the current floor as a topological prior to guide the exploration order. When exploring the initial floor, where no reference floor has been observed, planning relies only on the real graph ($\mathcal{G}_{\mathrm{real}}$), and the local layer directly schedules the observed TF clusters.

\subsection{Task-Zone Partition and Hypothesis Initialization}
\label{subsec:taskzone}

To derive macro-level planning units, we partition the real graph~$\mathcal{G}_{\mathrm{real}}$ of the already-explored floor (the \emph{reference floor}) into compact task zones. To ensure zone centers naturally align with open spaces rather than clustering near walls, we iteratively select the unassigned node with the largest horizontal distance to the nearest obstacle as a seed. Starting from this seed, a bounded Dijkstra expansion along valid edges extracts nodes into the current zone until the shortest-path distance exceeds a threshold~$r_z$. During expansion, previously covered nodes are reassigned if the new seed offers a shorter path. This procedure repeats until every node is covered; small localized zones are then absorbed into adjacent neighbors to reduce fragmentation. The resulting partition clusters spatially contiguous graph regions into distinct zones that respect structural barriers such as walls and doorways. For each zone~$Z_k$, the graph node closest to its spatial centroid is designated as the representative center~$c_k$, yielding the task-zone set $\mathcal{Z}_{\mathrm{ref}} = \{Z_1, \dots, Z_M\}$.

Upon transitioning to an unexplored \emph{target floor}, the real graph of the reference floor and its zone partitions are preserved and vertically translated by an estimated inter-floor height~$\Delta z$. This yields a hypothetical graph for the target floor, $\mathcal{G}_{\mathrm{hyp}}=(\mathcal{V}_{\mathrm{hyp}},\mathcal{E}_{\mathrm{hyp}})$, where each node is marked \emph{hypothetical}~($\mathcal{S}_{\mathrm{hyp}}$) and inherits its original task-zone label. This vertical projection imposes a structural prior onto the unknown target floor, strategically guiding early exploration before sufficient real observations accumulate.

\subsection{Incremental Maintenance via Promotion and Merge/Split}
\label{sec:inc_maintenance}

The projected hypothesis serves as an initial structural prior. As exploration proceeds, online graph expansion (Sec.~\ref{subsec:sparse_graph}) incrementally extends $\mathcal{G}_{\mathrm{real}}$, while hypothetical nodes ($\mathcal{S}_{\mathrm{hyp}}$) are confirmed or discarded based on actual observations. This continuous reconciliation dynamically maintains a unified \emph{hybrid graph} ($\mathcal{G}_{\mathrm{hybrid}} \triangleq \mathcal{G}_{\mathrm{real}} \cup \mathcal{G}_{\mathrm{hyp}}$) through node-level promotion and zone-level adaptation.

\subsubsection{Node Promotion}

A hypothetical node $\hat{v} \in \mathcal{V}_{\mathrm{hyp}}$ is promoted to a confirmed state ($\mathcal{S}_{\mathrm{hyp}} \to \mathcal{S}_{\mathrm{conf}}$) once the traversability map discovers a valid support surface near its projected column. During exploration, online graph expansion may also sample a real node in the same local support region. To avoid duplicates, such an overlapping node is merged into $\hat{v}$ by transferring its edges before deletion. If no nearby confirmed node exists, $\hat{v}$ simply transitions its state. Concurrently, incident hypothetical edges are validated against real terrain: collision-free connections become confirmed, while obstructed ones are pruned. If repeated measurements verify no traversable surface near $\hat{v}$, the hypothesis is permanently discarded.

\subsubsection{Task-Zone Split and Merge}

\begin{figure}[t]
    \begin{center}
    \includegraphics[width=\linewidth]{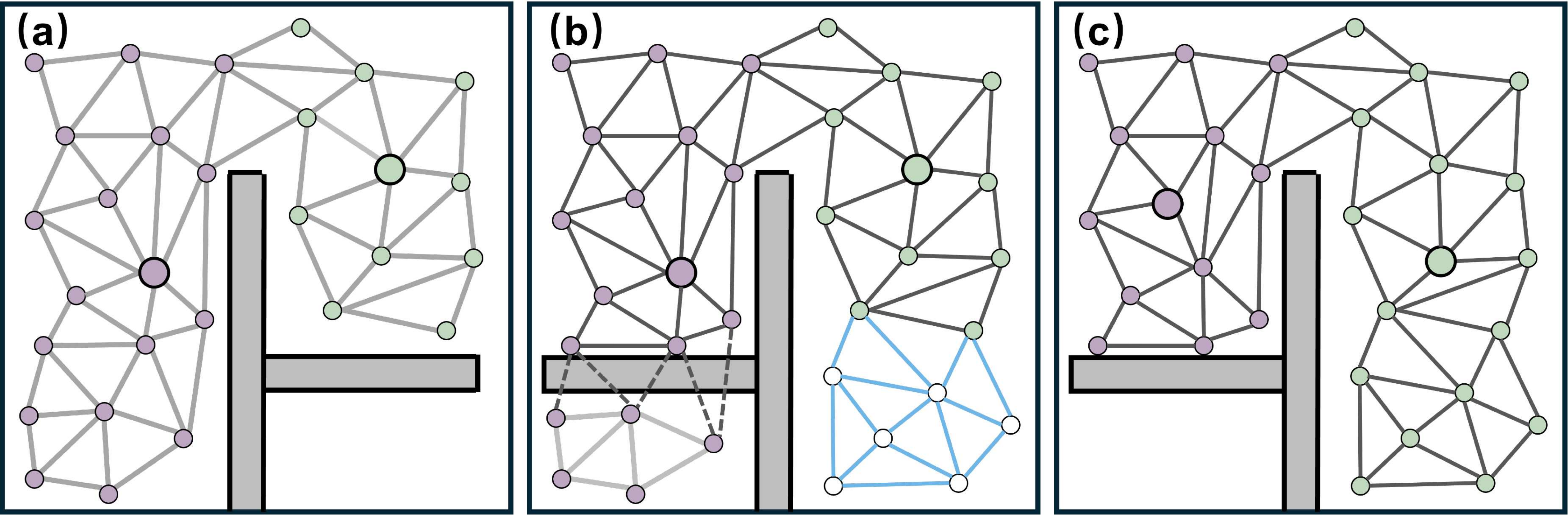}
    \end{center}
    \vspace{-0.3cm}
    \caption{\label{fig:task_zone_split}
    Incremental maintenance of task-zone partitions on the sparse graph. (a) Initial hypothesis projected from the reference floor, where large nodes represent zone centers ($c_k$). (b) Connectivity split and unassigned component extraction. Newly observed obstacles invalidate prior edges (dashed lines), triggering a zone split; newly sampled nodes form unassigned components (blue nodes). (c) Zone reassignment. Unassigned components merge into neighboring zones based on topological affinity, and representative centers are recomputed to align with updated spatial centroids.
    }
    % \vspace{-0.3cm}
\end{figure}

Target-floor task zones inherit their initial partitions from the reference floor, but actual layouts may vary due to closed doors or subdivided rooms. To align them with actual observations, they undergo a three-step structural adaptation (Fig.~\ref{fig:task_zone_split}):

\noindent\textbf{1) Connectivity split.}\; Newly discovered obstacles invalidate prior edges, necessitating internal connectivity re-evaluation for each zone~$Z_k$. We extract the connected component containing the representative center~$c_k$. If $c_k$ becomes isolated by obstacles, we default to the largest surviving component to designate a new center. All disconnected nodes, whether from $\mathcal{G}_{\mathrm{real}}$ or $\mathcal{G}_{\mathrm{hyp}}$, are detached and marked as unassigned.

\noindent\textbf{2) Unassigned component extraction.}\; The nodes detached during the split, along with any newly sampled unassigned nodes, are subsequently reorganized. By executing a breadth-first search over valid edges among these unassigned nodes, we extract topologically connected subsets $\{U_1, U_2, \dots\}$.

\noindent\textbf{3) Zone reassignment and merge.}\; For each subset~$U_j$, nearby existing zones are considered as merge candidates. Provided the target zone does not exceed its node capacity limit, $U_j$ is merged into a zone~$Z_k$ if they share sufficient cross-zone edges or if their centroids are spatially close. Otherwise, $U_j$ is initialized as an independent zone.

Finally, the representative center~$c_k$ of each modified zone is recomputed as the node nearest to its updated spatial centroid. This mechanism continuously corrects erroneous priors while integrating newly discovered regions.

\subsection{Global Guidance on the Hybrid Graph}
\label{sec:global_planning}

Each target-floor task zone is dynamically classified into one of three states $\sigma(Z_k) \in \{H, A, C\}$: \emph{hypothetical}~($H$) if it is exclusively supported by the initial projection; \emph{active}~($A$) if it contains nodes from $\mathcal{G}_{\mathrm{real}}$ paired with unresolved TF clusters; and \emph{completed}~($C$) if fully verified with no remaining exploration targets.

Unlike traditional methods confined to mapped regions, the macro planner formulates an open-loop TSP over all uncompleted zones, $\mathcal{Z}^{+}=\{Z_k \mid \sigma(Z_k) \neq C\}$. This joint scheduling of active and hypothetical zones yields a visiting sequence that spans the entire scene layout. The travel cost $D_{ij}$ between two corresponding representatives~$c_i$ and~$c_j$ is evaluated via an adaptive two-stage A* search. Whenever possible, the algorithm attempts to route exclusively through elements in the real state ($\mathcal{G}_{\mathrm{real}}$) to leverage reliable structures. When this is infeasible, such as when bridging disconnected mapped regions or navigating to hypothetical zones, the search adaptively expands to include hypothetical elements, routing over the full hybrid graph $\mathcal{G}_{\mathrm{hybrid}}$.

This adaptive routing yields a TSP tour that seamlessly integrates known free space with inferred geometric layouts. Execution enforces strict motion safety: when guiding toward a hypothetical region, the robot traverses only the real prefix of the path, observes the predicted boundary, incrementally refines the prior, and replans.

\subsection{Local Exploration Planning}
% \subsection{Heading-Aware Local Viewpoint Scheduling}
\label{sec:local_trajectory}

Unlike aerial platforms with omnidirectional agility, ground robots suffer significant execution delays from frequent reorientation. Consequently, local exploration must decide not only which frontier cluster to visit next, but also which viewpoint to use and how to execute the motion smoothly. Since penalizing heading changes depends on three consecutive viewpoints, standard TSP formulations based only on pairwise distance become inadequate. We therefore adopt a three-stage pipeline: Stage 1 determines the cluster visiting order, Stage 2 selects one heading-aware viewpoint for each chosen cluster, and Stage 3 generates a safe and smooth trajectory for execution.

\noindent\textbf{Stage 1: Cluster Sequencing.}\; Given a set of TF clusters, we solve a TSP over their centroids to determine the visiting order. When macro-layer guidance is available (Sec.~\ref{sec:global_planning}), we collect TF clusters from the first few task zones in the global sequence and use the representative center of the next unvisited task zone as a terminal anchor, denoted by $c_{\mathrm{anc}}$, to bias the local route toward the global plan. When exploring the first floor, where no macro guidance is available, the TSP directly schedules all available TF clusters without this anchor.

\noindent\textbf{Stage 2: Viewpoint Selection.}\; Along the resulting cluster sequence, we select exactly one viewpoint per cluster within a sliding window of size~$W$. Let $\mathcal{X}_m=\{\mathbf{x}_m^1,\dots,\mathbf{x}_m^{K_m}\}$ denote candidate viewpoints for the $m$-th cluster, filtered for visibility and safety. We find the optimal sequence via dynamic programming (DP), minimizing a window cost (Eq.~\eqref{eq:dp_cost}) that accounts for path lengths, heading changes ($\Delta \psi_m$), and global alignment. Since evaluating $\Delta \psi_m$ requires three consecutive waypoints, the DP state tracks the last two selected viewpoints, yielding a complexity of $O(W \cdot K^3)$, where $K$ is the maximum candidate count. In each replanning cycle, the first viewpoint in the optimal window is executed as the immediate local target.
\begingroup
\setlength{\abovedisplayskip}{2pt}
\setlength{\belowdisplayskip}{2pt}
\setlength{\jot}{0pt}
\begin{equation}
\begin{aligned}
J =\;& d(\mathbf{x}_0,\mathbf{x}_1^{k_1}) + \sum_{m=1}^{W-1} d(\mathbf{x}_m^{k_m},\mathbf{x}_{m+1}^{k_{m+1}}) \\[-2pt]
&+ \lambda_{\psi}\sum_{m=0}^{W-1}\Delta \psi_m + \lambda_a\, d(\mathbf{x}_W^{k_W},c_{\mathrm{anc}}),
\end{aligned}
\label{eq:dp_cost}
\end{equation}
\endgroup
\noindent where $\mathbf{x}_0$ is the current robot pose, $d(\cdot,\cdot)$ is the A* path cost on the sparse graph, and $\lambda_{\psi}, \lambda_a \ge 0$ are tunable penalties.

\noindent\textbf{Stage 3: Safe Trajectory Generation.}\; Given the immediate local target, we compute a physically executable route by running A* through the free voxels above the traversable surface. The resulting discrete path is then optimized into a smooth, collision-free, and dynamically feasible B-spline trajectory for low-level tracking.

\section{Experiments}
\label{sec:experiments}

\subsection{Simulation Experiments}

\noindent\textbf{Implementation Details.}\; The proposed framework is evaluated in the Gazebo simulator, running on a desktop computer equipped with an Intel i7-13700HX CPU, an NVIDIA RTX 3060 GPU, and 32\,GB of RAM. To navigate complex multi-elevation terrains such as stairs, we employ a Unitree Go2 quadruped as the robotic platform, equipped with a Velodyne VLP-16 LiDAR ($360^{\circ} \times 30^{\circ}$ FOV) as the primary sensor.

\noindent\textbf{Simulation Environments.}\; To evaluate the proposed framework, three challenging multi-floor scenarios are designed: (1) \emph{Office}, a two-story workspace connected by a staircase, where corresponding rooms across floors have inconsistent door states and notably different local layouts in certain areas, challenging the system's ability to reconcile cross-floor structural discrepancies; (2) \emph{Maze}, a large-scale scene linked by an extremely gentle slope (about $5^{\circ}$), featuring significant cross-floor structural disparities and multiple complex intersections; and (3) \emph{Teaching Building}, a multi-floor structure comprising looped corridors, open hubs, and functional rooms, connected vertically by U-shaped return stairs.

\begin{figure}[t]
    \begin{center}
    \includegraphics[width=1.00\linewidth]{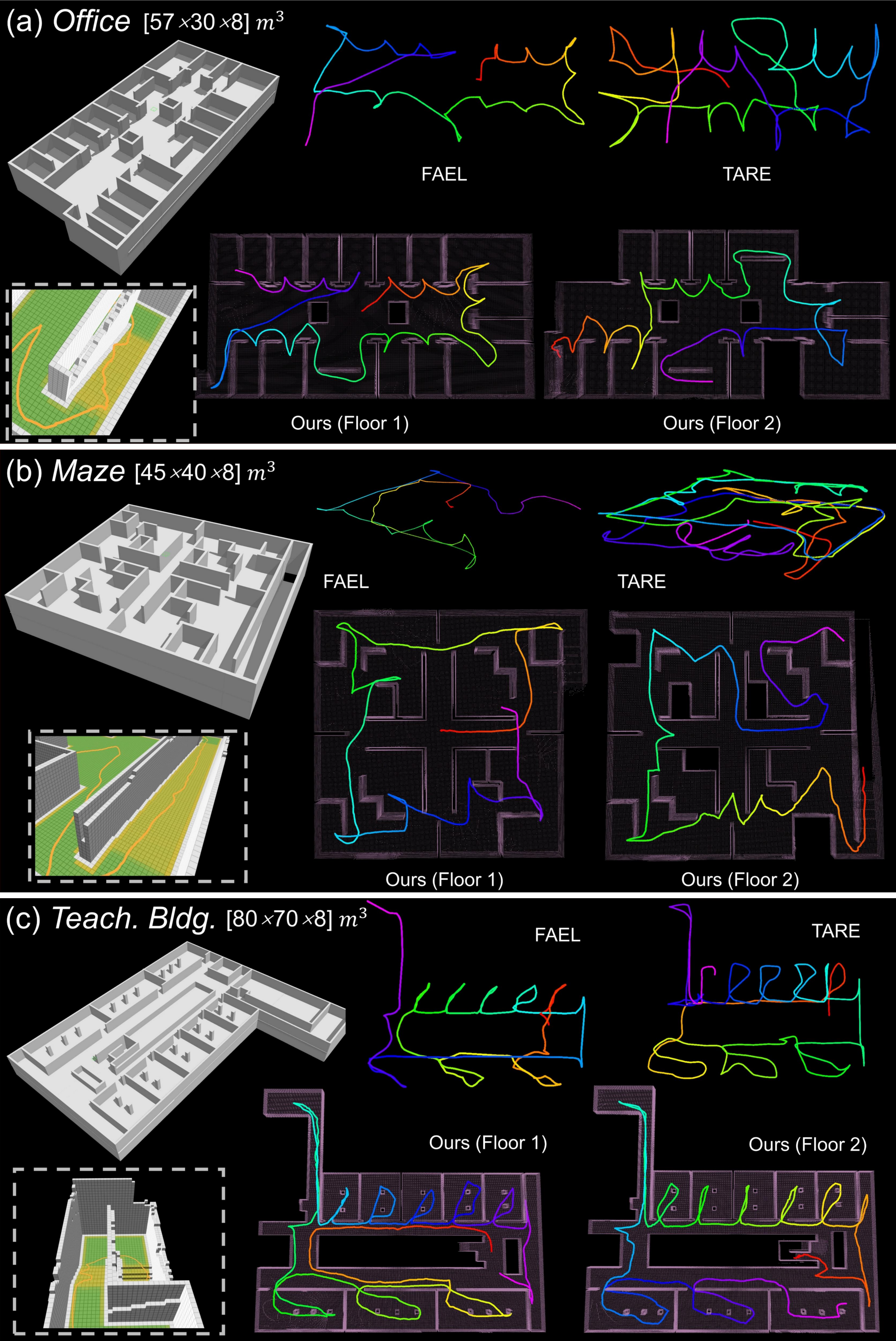}
    \end{center}
    \vspace{-0.3cm}
    \caption{\label{fig:simulation_experiments}
    Resulting maps from our method and trajectories of all methods in three simulation scenes. Red and purple denote the start and end of each trajectory, respectively; snapshots of cross-floor traversal are also shown.
    }
    % \vspace{-0.3cm}
\end{figure}

\noindent\textbf{Comparisons and Analysis.}\; We benchmark the proposed framework against two representative ground exploration baselines: TARE~\cite{cao2021tare} and FAEL~\cite{huang2023fael}. Performance is evaluated using three primary metrics: total exploration time, trajectory length, and mapping completeness. Unlike conventional evaluations that report only the explored occupied volume, we additionally measure the volume of reconstructed traversable voxels to better quantify coverage of the navigable structure. To ensure an equitable comparison, all methods operate under identical speed limits, sensor setups, mapping parameters, and a unified statistics module. Each configuration is evaluated across five independent runs. Quantitative results are summarized in Tables~\ref{tab:sim_benchmark} and~\ref{tab:per_floor}, while Fig.~\ref{fig:simulation_experiments} shows the resulting maps and representative trajectories.

\begin{table}[t]
\centering
\caption{Results of Simulations in Three Multi-Elevation \\ Environments}
\label{tab:sim_benchmark}
\vspace{-0.1cm}
\renewcommand{\arraystretch}{1.2}
\setlength{\tabcolsep}{2.5pt}
\resizebox{\columnwidth}{!}{
\begin{tabular}{c|c|c|c|cc|cc|cc|cc}
\hline
\multirow{2}{*}{\textbf{Scene}} & \multirow{2}{*}{\textbf{Method}} & \multirow{2}{*}{\textbf{Succ.}} & \multirow{2}{*}{\textbf{Floor}} & \multicolumn{2}{c|}{\makecell{\textbf{Exp.}\\\textbf{Time (s)}}} & \multicolumn{2}{c|}{\makecell{\textbf{Traj.}\\\textbf{Dist. (m)}}} & \multicolumn{2}{c|}{\makecell{\textbf{Trav.}\\\textbf{Vol. (m$^3$)}}} & \multicolumn{2}{c}{\makecell{\textbf{Occ.}\\\textbf{Vol. (m$^3$)}}} \\
\cline{5-12}
& & & & \textbf{Avg} & \textbf{Std} & \textbf{Avg} & \textbf{Std} & \textbf{Avg} & \textbf{Std} & \textbf{Avg} & \textbf{Std} \\
\hline
\multirow{3}{*}{\textit{Office}}
& Ours & Yes & 1+2 & \textbf{530} & 17 & 470 & 16 & \textbf{475} & 1 & \textbf{1223} & 9 \\
& TARE & No  & 1   & 594 & 48 & 449 & 27 & 257 & 3 & 661  & 14 \\
& FAEL & No  & 1   & 347 & 65 & 173 & 3  & 228 & 3 & 590  & 6 \\
\hline
\multirow{3}{*}{\textit{Maze}}
& Ours & Yes & 1+2 & \textbf{542} & 36 & \textbf{492} & 35 & \textbf{514} & 1 & \textbf{1273} & 12 \\
& TARE & Yes & 1+2 & 1027& 41 & 826 & 38 & 501 & 2 & 1270 & 16 \\
& FAEL & No  & 1   & 397 & 17 & 256 & 6  & 255 & 1 & 603  & 7 \\
\hline
\multirow{3}{*}{\makecell{\textit{Teach.}\\\textit{Bldg.}}}
& Ours & Yes & 1+2 & \textbf{1253}& 25 & 1199& 16 & \textbf{992} & 1 & \textbf{2435} & 17 \\
& TARE & No  & 1   & 922 & 62 & 788 & 44 & 487 & 1 & 1210 & 21 \\
& FAEL & No  & 1   & 788 & 62 & 553 & 47 & 474 & 13& 1157 & 32 \\
\hline
\end{tabular}
}
\begin{tablenotes}
 \scriptsize
 \item[] $\dagger$ \textit{Succ.} indicates whether the method successfully explores all floors. \textit{Floor} specifies the successfully explored building stories.
\end{tablenotes} 
% \vspace{-0.5cm}
\end{table}

\noindent\textit{(1) Multi-elevation exploration capability.}\;
FAEL remains on a single floor across all scenarios. This is because it projects 3D sensor data onto a 2D traversability grid, where regions with significant elevation changes, such as staircases, are flagged as occupied due to excessive height variance, and both viewpoint sampling and frontier detection are confined to this plane. TARE can reach the upper floor in \emph{Maze}, but only through the continuous gentle ramp. Its height-bounded connectivity heuristic links adjacent viewpoints only when their elevation difference falls below a fixed threshold, which prevents traversal over steep stair transitions.

\noindent\textit{(2) Exploration efficiency.}\;
In \emph{Office} and \emph{Teaching Building}, the baselines fail to reach the second floor. While this highlights our method's multi-floor capability, it restricts efficiency comparisons to a single-floor basis. We therefore compare our first-floor completion times in Table~\ref{tab:per_floor} against the baselines' total runtimes in Table~\ref{tab:sim_benchmark}. In both scenes, our method finishes the first floor faster than the baselines terminate, demonstrating high exploration efficiency even in single-level environments. TARE suffers from repeated target invalidation and reactivation, which, combined with the limitations of its hierarchical planner, leads to incomplete room coverage and redundant backtracking, while FAEL often misses valid traversable regions and revisits explored areas.

A direct multi-floor comparison is available in \emph{Maze}, where TARE also reaches the upper floor via the ramp. Here, our framework reduces total exploration time by 47\% (542\,s vs.\ 1027\,s for TARE) and path length by 40\% (492\,m vs.\ 826\,m for TARE). These gains mainly come from stable exploration targets on the maintained reachable graph and structural priors that provide global guidance across floors. By contrast, TARE frequently shuttles between floors without consistent guidance, resulting in substantial redundant motion.
\begin{table}[h]
\centering
\caption{Per-Floor Statistics of the Proposed Framework}
\label{tab:per_floor}
\vspace{-0.1cm}
\renewcommand{\arraystretch}{1.2}
\setlength{\tabcolsep}{5.0pt}
\scriptsize
\begin{tabular}{c|c|c|c|c|c}
\hline
& \makecell{\textbf{1st Floor}\\\textbf{(s)}} & \makecell{\textbf{2nd Floor}\\\textbf{(s)}} & \makecell{\textbf{Reduction}} & \makecell{\textbf{Trav.}\\\textbf{Vol. (m$^3$)}} & \makecell{\textbf{Occ.}\\\textbf{Vol. (m$^3$)}} \\
\hline
\textit{Office}       & 271.4 & 224.0 & $17.5\%$ & 256.8 & 664.4 \\
\textit{Maze}         & 252.8 & 229.6 & $9.2\%$  & 258.4 & 628.1 \\
\textit{Teach. Bldg.} & 658.0 & 547.6 & $16.8\%$ & 489.4 & 1211.7 \\
\hline
\end{tabular}
\begin{tablenotes}
 \scriptsize
 \item[] $\dagger$ Per-floor times denote complete exploration within each floor, excluding cross-floor transitions. \textit{Reduction} denotes the time saved on the 2nd floor relative to the 1st floor. Volumetric metrics indicate the structures reconstructed only on the 1st floor.
\end{tablenotes}
% \vspace{-0.2cm}
\end{table}

\noindent\textit{(3) Mapping completeness.}\;
Our method achieves the highest mapping completeness in all scenarios. TARE sometimes attains comparable reconstructed volumes, but this is largely because its inefficient routing produces longer trajectories and repeated revisits, allowing the robot to passively scan more areas during prolonged operation. FAEL shows lower completeness because its 2D traversability representation limits its ability to capture out-of-plane structures.

\noindent\textit{(4) Cross-floor guidance from structural priors.}\;
As evidenced by Table~\ref{tab:per_floor}, our framework explores the target floor faster than the reference floor across all scenarios. This consistent $9.2\%$ to $17.5\%$ reduction in exploration time indicates that the projected structural prior provides effective layout-level guidance. By leveraging this prior, the hierarchical planner forms an efficient initial routing schedule on the target floor before dense observations accumulate.

\subsection{Ablation Experiments}
\label{sec:ablation_experiments}

To isolate the contribution of each proposed component, we conduct two groups of ablation studies. Group~A evaluates the reachable graph and frontier detection mechanism on a single floor, while Group~B evaluates the cross-floor structural prior on the target floor of a multi-floor scenario.

\subsubsection{\textbf{Group A: Reachable Graph and Frontier Detection}}
\label{sec:ablation_topology_frontier}

We compare five configurations on the first floor of the \emph{Maze} environment:
\textbf{Full}: the complete proposed framework.
\textbf{w/o tentative}: the tentative mechanism is disabled; the graph is restricted strictly to $\mathcal{C}_{\mathrm{reach}}$.
\textbf{w/o surface criterion}: disables $C_{\mathrm{sur}}$, relying solely on topological extensibility.
\textbf{w/o expansion criterion}: disables $C_{\mathrm{exp}}$, relying solely on surface incompleteness.
\textbf{Boundary frontier}: replaces our topological frontiers with frontiers extracted from the geometric boundary of $\mathcal{C}_{\mathrm{reach}}$; all other steps follow the proposed pipeline.

\begin{table}[h]
\centering
\caption{Ablation on Reachable Graph and Frontier Detection}
\label{tab:ablation_a}
\vspace{-0.1cm}
\renewcommand{\arraystretch}{1.2}
\setlength{\tabcolsep}{3pt}
\scriptsize
\begin{tabular}{l|c|c|c}
\hline
\textbf{Configuration} & \textbf{Exp.\ Time (s)} & \textbf{Traj.\ Dist.\ (m)} & \textbf{Occ.\ Vol.\ (m$^3$)} \\
\hline
Full (Ours)             & \textbf{253} & \textbf{221} & \textbf{628} \\
w/o tentative           & 334 & 295 & 617 \\
w/o surface crit.       & 249 & 227 & 608 \\
w/o expansion crit.     & 250 & 217 & 595 \\
Boundary frontier       & 326 & 294 & 624 \\
\hline
\end{tabular}
% \begin{tablenotes}
%  \scriptsize
%  \item[]$\dagger$ Evaluated on the 1st floor of \textit{Maze} (5 runs).
% \end{tablenotes}
% \vspace{-0.3cm}
\end{table}

As shown in Table~\ref{tab:ablation_a}, removing the tentative mechanism restricts graph growth strictly to fully confirmed regions. This prevents the graph from interfacing with the true unknown, which delays shortcut discovery, forces conservative planning, and ultimately degrades efficiency. Disabling either criterion misses valid targets; the deceptively shorter exploration time is merely an artifact of premature termination, evidenced by lower reconstructed volumes. Finally, the boundary-frontier baseline yields fragmented targets, resulting in a clear drop in efficiency compared to our approach.

\subsubsection{\textbf{Group B: Cross-Floor Structural Prior}}
\label{sec:ablation_prior}

We evaluate three configurations on the second floor of the \emph{Maze} environment.
\textbf{Full}: the complete method with hypothesis initialization and incremental maintenance.
\textbf{w/o prior}: the cross-floor prior is disabled; the target floor is explored without any hypothetical graph, relying only on the incrementally built $\mathcal{G}_{\mathrm{real}}$.
\textbf{w/o incremental maintenance}: the hypothetical graph is initialized and its nodes can be confirmed, but connectivity splitting and zone reassignment are disabled, leaving the initial macro-level hypothesis fixed during exploration.

\begin{table}[h]
\centering
\caption{Ablation on Cross-Floor Structural Prior}
\label{tab:ablation_b}
\vspace{-0.1cm}
\renewcommand{\arraystretch}{1.2}
\setlength{\tabcolsep}{3pt}
\scriptsize
\begin{tabular}{l|c|c|c}
\hline
\textbf{Configuration} & \textbf{Exp.\ Time (s)} & \textbf{Traj.\ Dist.\ (m)} & \textbf{Occ.\ Vol.\ (m$^3$)} \\
\hline
Full (Ours)                & \textbf{230} & \textbf{217} & \textbf{574} \\
w/o prior                  & 272 & 253 & 540 \\
w/o incr.\ maint.          & 282 & 261 & 519 \\
\hline
\end{tabular}
% \vspace{-0.2cm}
\end{table}

As shown in Table~\ref{tab:ablation_b}, without the structural prior, the planner lacks global layout guidance on the target floor and must discover the environment reactively, increasing exploration time and path length by 18\% and 17\%, respectively. When the prior is provided but not incrementally maintained, stale hypotheses persist despite conflicting observations. These uncorrected discrepancies misdirect global routing, further increasing time and path length by 23\% and 20\%.

\subsection{Real-World Experiments}
\label{sec:real_world_experiments}

To validate deployment performance, we conduct experiments using a Unitree Go2 quadruped equipped with a Livox Mid-360 LiDAR and an Intel NUC12 (i7-1260P CPU, 32\,GB RAM). All modules run fully onboard in real time, with state estimation provided by a modified FAST-LIO2~\cite{xu2022fast}.

The testing environment spans two adjacent floors of a campus building, featuring loop-shaped corridors, interconnected lobbies, and U-shaped return staircases. Although the two floors share a similar overall footprint, their local layouts differ in several areas~(Fig.~\ref{fig:head_img}). This setup evaluates practical exploration efficiency and the system's ability to reconcile mismatched cross-floor structural priors. Without global layout guidance, first-floor exploration took 321\,s and traversed long corridors to revisit missed areas. On the second floor, the projected structural prior reduced exploration time to 226\,s, a $29.6\%$ reduction. Two additional deployments highlight the framework's versatility: a large-scale courtyard building scene and an underground garage connected to the surface by a steep ramp with a $90^{\circ}$ turn, which can be challenging for conventional 2D ground exploration algorithms (Fig.~\ref{fig:real_world_experiments}). Further details are available in the supplementary video.
\begin{figure}[t]
    \begin{center}
    % \includegraphics[width=\linewidth]{img_compress/tb1.png}
    % \vspace{0.05cm}
    % \includegraphics[width=\linewidth]{img_compress/garage.png}
    \includegraphics[width=\linewidth]{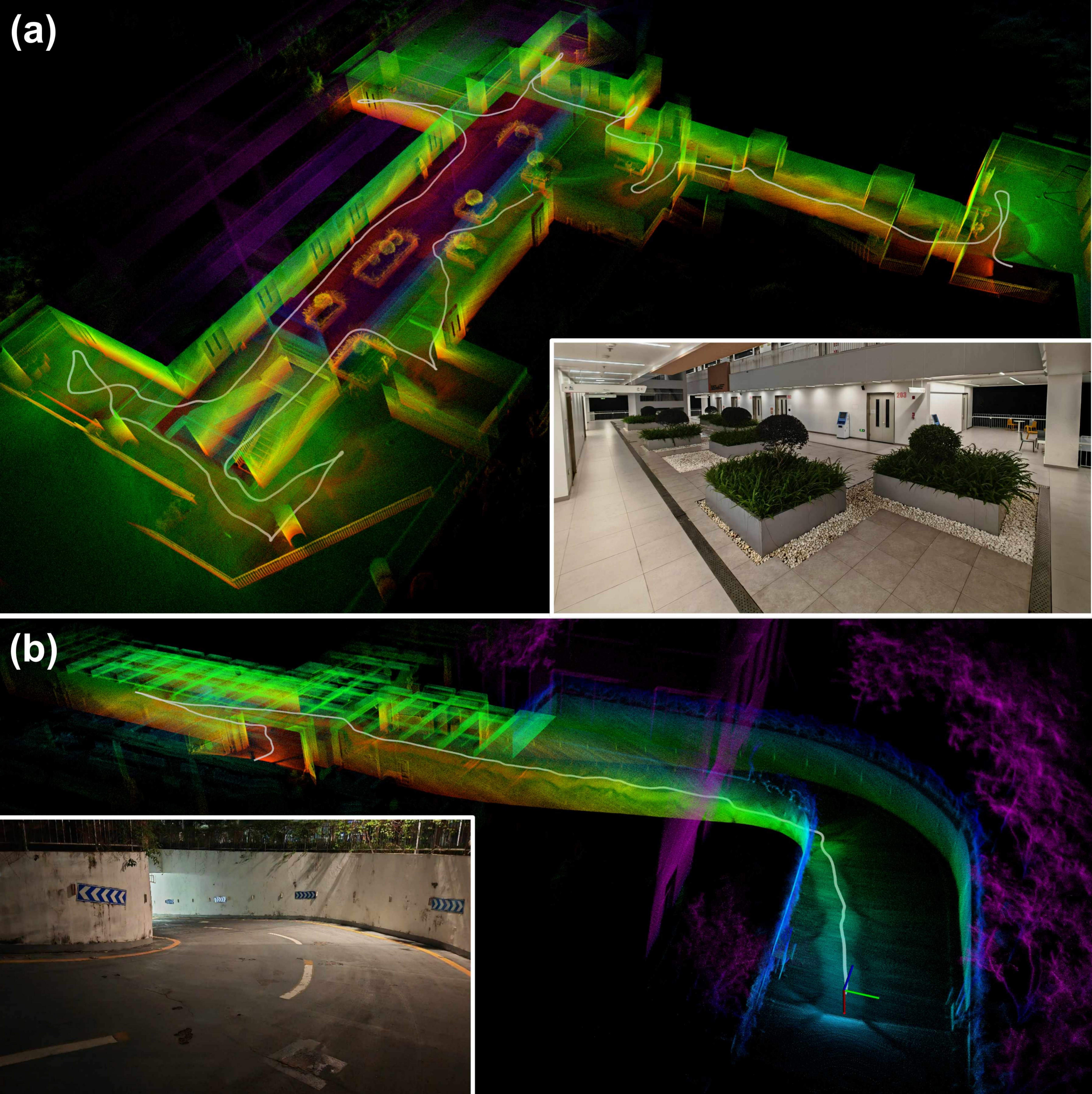}
    \end{center}
    \vspace{-0.3cm}
    \caption{\label{fig:real_world_experiments}
    Real-world experiments in (a) a large courtyard building scene and (b) an underground garage featuring a steep ramp with a $90^{\circ}$ turn. Each scene shows the environment photograph, final point cloud, and complete trajectory.
    }
    % \vspace{-0.3cm}
\end{figure}

\section{Conclusions}
\label{sec:conclusions}

This letter presented an exploration framework for ground robots in multi-floor environments. By incrementally maintaining a traversability-aware reachable graph with tentative connectivity, our method generates stable, reachable exploration targets. Cross-floor structural priors further guide hierarchical planning across floors. Simulations and real-world quadruped experiments show improved exploration efficiency and mapping completeness. A limitation is the reliance of the macro layer on inter-floor structural regularity; when adjacent floors differ substantially, planning degrades to conventional exploration on the observed graph. Future work will address scenarios with weak inter-floor structural correspondence.

\begingroup
\footnotesize
\def\IEEEbibitemsep{0pt}
\bibliography{zzw} 

@string{icra = {Proc. of the {IEEE} Intl. Conf. on Robot. and Autom. ({ICRA})}}

@string{iros = {Proc. of the {IEEE/RSJ} Intl. Conf. on Intell. Robots and Syst.({IROS})}}

@article{zhou2021fuel,
  title={Fuel: Fast uav exploration using incremental frontier structure and hierarchical planning},
  author={Zhou, Boyu and Zhang, Yichen and Chen, Xinyi and Shen, Shaojie},
  journal={IEEE Robotics and Automation Letters},
  volume={6},
  number={2},
  pages={779--786},
  year={2021},
  publisher={IEEE}
}

@inproceedings{bircher2016receding,
  title={Receding horizon" next-best-view" planner for 3d exploration},
  author={Bircher, Andreas and Kamel, Mina and Alexis, Kostas and Oleynikova, Helen and Siegwart, Roland},
  booktitle={2016 IEEE international conference on robotics and automation (ICRA)},
  pages={1462--1468},
  year={2016},
  organization={IEEE}
}

@article{zhou2023racer,
  title={Racer: Rapid collaborative exploration with a decentralized multi-uav system},
  author={Zhou, Boyu and Xu, Hao and Shen, Shaojie},
  journal={IEEE Transactions on Robotics},
  year={2023},
  publisher={IEEE}
}

@inproceedings{cao2021tare,
  title={TARE: A Hierarchical Framework for Efficiently Exploring Complex 3D Environments.},
  author={Cao, Chao and Zhu, Hongbiao and Choset, Howie and Zhang, Ji},
  booktitle={Robotics: Science and Systems},
  volume={5},
  year={2021}
}

@article{xu2022fast,
  title={Fast-lio2: Fast direct lidar-inertial odometry},
  author={Xu, Wei and Cai, Yixi and He, Dongjiao and Lin, Jiarong and Zhang, Fu},
  journal={IEEE Transactions on Robotics},
  volume={38},
  number={4},
  pages={2053--2073},
  year={2022},
  publisher={IEEE}
}

@article{zhang2024falcon,
  title={Falcon: Fast autonomous aerial exploration using coverage path guidance},
  author={Zhang, Yichen and Chen, Xinyi and Feng, Chen and Zhou, Boyu and Shen, Shaojie},
  journal={IEEE Transactions on Robotics},
  year={2024},
  publisher={IEEE}
}

@inproceedings{yamauchi1998frontier,
  title={Frontier-based exploration using multiple robots},
  author={Yamauchi, Brian},
  booktitle={Proceedings of the second international conference on Autonomous agents},
  pages={47--53},
  year={1998}
}

@article{krusi2017driving,
  title={Driving on Point Clouds: Motion Planning, Trajectory Optimization, and Terrain Assessment in Generic Nonplanar Environments},
  author={Kr{\"u}si, Philipp and Furgale, Paul and Bosse, Michael and Siegwart, Roland},
  journal={Journal of Field Robotics},
  volume={34},
  number={5},
  pages={940--984},
  year={2017},
  doi={10.1002/rob.21700}
}

@inproceedings{azpurua2021terrain,
  title={Three-dimensional Terrain Aware Autonomous Exploration for Subterranean and Confined Spaces},
  author={Azpurua, Hector and Campos, Mario F. M. and Macharet, Douglas G.},
  booktitle={2021 IEEE International Conference on Robotics and Automation (ICRA)},
  pages={2443--2449},
  year={2021},
  organization={IEEE},
  doi={10.1109/ICRA48506.2021.9561099}
}

@inproceedings{miki2022elevation,
  title={Elevation Mapping for Locomotion and Navigation using GPU},
  author={Miki, Takahiro and Wellhausen, Lorenz and Grandia, Ruben and Jenelten, Fabian and Homberger, Timon and Hutter, Marco},
  booktitle={2022 IEEE/RSJ International Conference on Intelligent Robots and Systems (IROS)},
  pages={2273--2280},
  year={2022},
  organization={IEEE},
  doi={10.1109/IROS47612.2022.9981507}
}

@inproceedings{chen2025lite,
  title={LITE: A Learning-Integrated Topological Explorer for Multi-Floor Indoor Environments},
  author={Chen, Junhao and Zhang, Zhen and Zhu, Chengrui and Hou, Xiaojun and Hu, Tianyang and Wu, Huifeng and Liu, Yong},
  booktitle={2025 IEEE/RSJ International Conference on Intelligent Robots and Systems (IROS)},
  pages={9533--9540},
  year={2025},
  organization={IEEE},
  doi={10.1109/IROS60139.2025.11246317}
}

@article{li2025multilevel,
  title={Real-Time Multilevel Terrain-Aware Path Planning for Ground Mobile Robots in Large-Scale Rough Terrains},
  author={Li, Yuxiang and Chen, Kun and Wang, Yifei and Zhang, Weifan and Wang, Jiancheng and Chen, Haoyao and Liu, Yunhui},
  journal={IEEE Transactions on Robotics},
  volume={41},
  pages={4159--4179},
  year={2025},
  doi={10.1109/TRO.2025.3577015}
}

@article{lee2025trg,
  title={TRG-Planner: Traversal Risk Graph-Based Path Planning in Unstructured Environments for Safe and Efficient Navigation},
  author={Lee, Dongkyu and Nahrendra, I Made Aswin and Oh, Minho and Yu, Byeongho and Myung, Hyun},
  journal={IEEE Robotics and Automation Letters},
  volume={10},
  number={2},
  pages={1736--1743},
  year={2025},
  publisher={IEEE},
  doi={10.1109/LRA.2024.3524912}
}

@inproceedings{hou2025fsgp,
  title={Real-time Spatial-temporal Traversability Assessment via Feature-based Sparse Gaussian Process},
  author={Tan, Senming and Hou, Zhenyu and Zhang, Zhihao and Xu, Long and Zhang, Mengke and He, Zhaoqi and Xu, Chao and Gao, Fei and Cao, Yanjun},
  booktitle={2025 IEEE/RSJ International Conference on Intelligent Robots and Systems (IROS)},
  pages={17533--17540},
  year={2025},
  organization={IEEE},
  doi={10.1109/IROS60139.2025.11246696}
}

@article{oh2024trip,
  title={TRIP: Terrain Traversability Mapping With Risk-Aware Prediction for Enhanced Online Quadrupedal Robot Navigation},
  author={Oh, Minho and Yu, Byeongho and Nahrendra, I Made Aswin and Jang, Seoyeon and Lee, Hyeonwoo and Lee, Dongkyu and Lee, Seungjae and Kim, Yeeun and Christiansen, Marsim Kevin and Lim, Hyungtae and Myung, Hyun},
  journal={arXiv preprint arXiv:2411.17134},
  year={2024},
  doi={10.48550/arXiv.2411.17134}
}

@inproceedings{jian2022putn,
  title={PUTN: A Plane-fitting based Uneven Terrain Navigation Framework},
  author={Jian, Zhuozhu and Lu, Zihong and Zhou, Xiao and Lan, Bin and Xiao, Anxing and Wang, Xueqian and Liang, Bin},
  booktitle={2022 IEEE/RSJ International Conference on Intelligent Robots and Systems (IROS)},
  pages={7160--7166},
  year={2022},
  organization={IEEE},
  doi={10.1109/IROS47612.2022.9981038}
}

@article{yang2025tomography,
  title={Efficient Global Navigational Planning in 3-D Structures Based on Point Cloud Tomography},
  author={Yang, Bowen and Cheng, Jie and Xue, Bohuan and Jiao, Jianhao and Liu, Ming},
  journal={IEEE/ASME Transactions on Mechatronics},
  volume={30},
  number={1},
  year={2025},
  doi={10.1109/TMECH.2024.3396001}
}

@inproceedings{zhang2025planes,
  title={Efficient Trajectory Generation Based on Traversable Planes in 3D Complex Architectural Spaces},
  author={Zhang, Mengke and Tian, Zhihao and Xia, Yaoguang and Xu, Chao and Gao, Fei and Cao, Yanjun},
  booktitle={2025 IEEE International Conference on Robotics and Automation (ICRA)},
  pages={14513--14519},
  year={2025},
  organization={IEEE},
  doi={10.1109/ICRA55743.2025.11128727}
}

@article{zhang2025ddr,
  title={Universal Trajectory Optimization Framework for Differential Drive Robot Class},
  author={Zhang, Mengke and Chen, Nanhe and Wang, Hu and Qiu, Jianxiong and Han, Zhichao and Ren, Qiuyu and Xu, Chao and Gao, Fei and Cao, Yanjun},
  journal={IEEE Transactions on Automation Science and Engineering},
  volume={22},
  pages={13030--13045},
  year={2025},
  publisher={IEEE},
  doi={10.1109/TASE.2025.3550676}
}

@article{elfes1989using,
  title={Using Occupancy Grids for Mobile Robot Perception and Navigation},
  author={Elfes, Alberto},
  journal={Computer},
  volume={22},
  number={6},
  pages={46--57},
  year={1989},
  doi={10.1109/2.30720}
}

@article{huang2023fael,
  title={FAEL: Fast Autonomous Exploration for Large-scale Environments With a Mobile Robot},
  author={Huang, Junlong and Zhou, Boyu and Fan, Zhengping and Zhu, Yilin and Jie, Yingrui and Li, Longwei and Cheng, Hui},
  journal={IEEE Robotics and Automation Letters},
  volume={8},
  number={3},
  pages={1667--1674},
  year={2023},
  doi={10.1109/LRA.2023.3236573}
}

@inproceedings{liu2024sfre,
  title={SFRE: Safe and Fast Robotic Exploration for 3D Uneven Terrains},
  author={Liu, Shengkai and Wang, Runhua and Bi, Qingchen and Wen, Guanghui and Zhang, Xuebo},
  booktitle={2024 IEEE/ASME International Conference on Advanced Intelligent Mechatronics (AIM)},
  pages={1--6},
  year={2024},
  organization={IEEE},
  doi={10.1109/AIM55361.2024.10637037}
}

@article{jia2025multiresolution,
  title={Unlocking Full Exploration Potential of Ground Robots by Multiresolution Topological Mapping},
  author={Jia, Yinghao and Wang, Changhong and Tang, Wei and Wang, Zhigang and Sun, Zhe},
  journal={IEEE Transactions on Industrial Informatics},
  volume={21},
  number={8},
  pages={6387--6397},
  year={2025},
  doi={10.1109/TII.2025.3563556}
}

@inproceedings{xu2024pare,
  title={PARE: A Plane-Assisted Autonomous Robot Exploration Framework in Unknown and Uneven Terrain},
  author={Xu, Pu and Bai, Zhaoqiang and Liu, Haoming and Fang, Zheng},
  booktitle={2024 IEEE/RSJ International Conference on Intelligent Robots and Systems (IROS)},
  pages={11707--11714},
  year={2024},
  organization={IEEE},
  doi={10.1109/IROS58592.2024.10802459}
}

@article{wang2020activemapping,
  title={Actively Mapping Industrial Structures with Information Gain-Based Planning on a Quadruped Robot},
  author={Wang, Yiduo and Ramezani, Milad and Fallon, Maurice},
  journal={IEEE Robotics and Automation Letters},
  volume={5},
  number={4},
  pages={7535--7542},
  year={2020},
  doi={10.1109/LRA.2020.3004451}
}

@article{brandao2020gaitmesh,
  title={GaitMesh: Controller-aware navigation meshes for long-range legged locomotion planning in multi-layered environments},
  author={Brandao, Martim and Aladag, Omer Burak and Havoutis, Ioannis},
  journal={IEEE Robotics and Automation Letters},
  volume={5},
  number={2},
  pages={3596--3603},
  year={2020},
  publisher={IEEE}
}
\endgroup

\end{document}